\documentclass{article}
\usepackage{ijcai20}

\usepackage{times}

\usepackage{soul}
\usepackage{url}
\usepackage[hidelinks]{hyperref}
\usepackage{graphicx}
\usepackage{subfigure}
\usepackage[utf8]{inputenc}
\usepackage[small]{caption}
\usepackage{amsmath}
\usepackage{amssymb}
\usepackage{booktabs}
\usepackage{algorithm}
\usepackage{algorithmic}
\usepackage{latexsym}
\usepackage{multirow}
\title{ACFD: Asymmetric Cartoon Face Detector}

\author{
Bin Zhang$^{12}$\footnote{Equal contribution. This work was down when Bin Zhang was an intern at Tencent Youtu Lab.}\and
Jian Li$^{1\ast}$\and
Yabiao Wang$^1$\and
Zhipeng Cui$^1$\and \\
Yili Xia$^2$\footnote{Contact Author}\and
Chengjie Wang$^1$\and
Jilin Li$^1$\and
Feiyue Huang$^1$
\\
\affiliations
$^1$Youtu Lab, Tencent\\
$^2$School of Information Science and Engineering, Southeast University\\
\emails
\{z-bingo, yili\_xia\}@seu.edu.cn \\
\{swordli, caseywang, zhipengcui, jasoncjwang, jerolinli, garyhuang\}.tencent.com
}

\begin{document}
\maketitle

\begin{abstract}
Cartoon face detection is a more challenging task than human face detection due to many difficult scenarios is involved. Aiming at the characteristics of cartoon faces, such as huge differences within the intra-faces, in this paper, we propose an asymmetric cartoon face detector, named ACFD. Specifically, it consists of the following modules: a novel backbone VoVNetV3 comprised of several asymmetric one-shot aggregation modules (AOSA), asymmetric bi-directional feature pyramid network (ABi-FPN), dynamic anchor match strategy (DAM) and the corresponding margin binary classification loss (MBC). In particular, to generate features with diverse receptive fields, multi-scale pyramid features are extracted by VoVNetV3, and then fused and enhanced simultaneously by ABi-FPN for handling the faces in some extreme poses and have disparate aspect ratios. Besides, DAM is used to match enough high-quality anchors for each face, and MBC is for the strong power of discrimination. With the effectiveness of these modules, our ACFD achieves the 1st place on the detection track of 2020 iCartoon Face Challenge under the constraints of model size 200MB, inference time 50ms per image, and without any pretrained models.

\end{abstract}

\section{Introduction}
Face detection is a long-standing and essential task for many downstream applications, such as face alignment, face recognition and face tracking. In order to detect face effectively and efficiently, many detection pipelines \cite{zhang2017s3fd,tang2018pyramidbox,chi2019srn,li2019dsfd,zhang2020refineface} have been proposed and achieved better and better performance on the challenging face detection benchmarks, such as WIDER Face \cite{yang2016wider}. In particular, on the most challenging dataset WIDER Face, the average precision (AP) has been improved from the original $40\%$ to more than $90\%$, and there are still many researchers are working on solving the remaining difficulties.

With the development of multimedia technology, more and more cartoon faces appear in the animations, comics, and even appear on the social media platforms as the personal profile. Therefore, the excellent detector for cartoon faces is also particularly important. Although the recent face detectors could handle most of the human faces, they cannot detect the cartoon faces accurately with many missing and false-positive results. That is because many scenarios in cartoons are more difficult than the real world, for instance, different faces may perform completely different characteristics, many faces are very similar to the negative samples (the bodies, background, etc.) and a considerable part of faces have the disparate aspect ratio of more than 3 or less than 1/3, which is almost not involved in human faces.

In this way, a large-scale and challenging cartoon person dataset is beneficial for designing an effective and robust deep learning based-approach. Several works \cite{gong2020autotoon,jha2018bringing,wu2019landmark} transfer the human faces to cartoons directly by utilizing generative adversarial networks due to the lack of cartoon face dataset, but there is still a big gap. Recently, a manually annotated cartoon face dataset is proposed by iQIYI \cite{li2019icartoonface} that contains 50000 images for detection tasks and more than 380000 images for face recognition, promoting the development of the community.

\begin{figure*}[!htbp]
    \centering
    \includegraphics[width=0.8\linewidth]{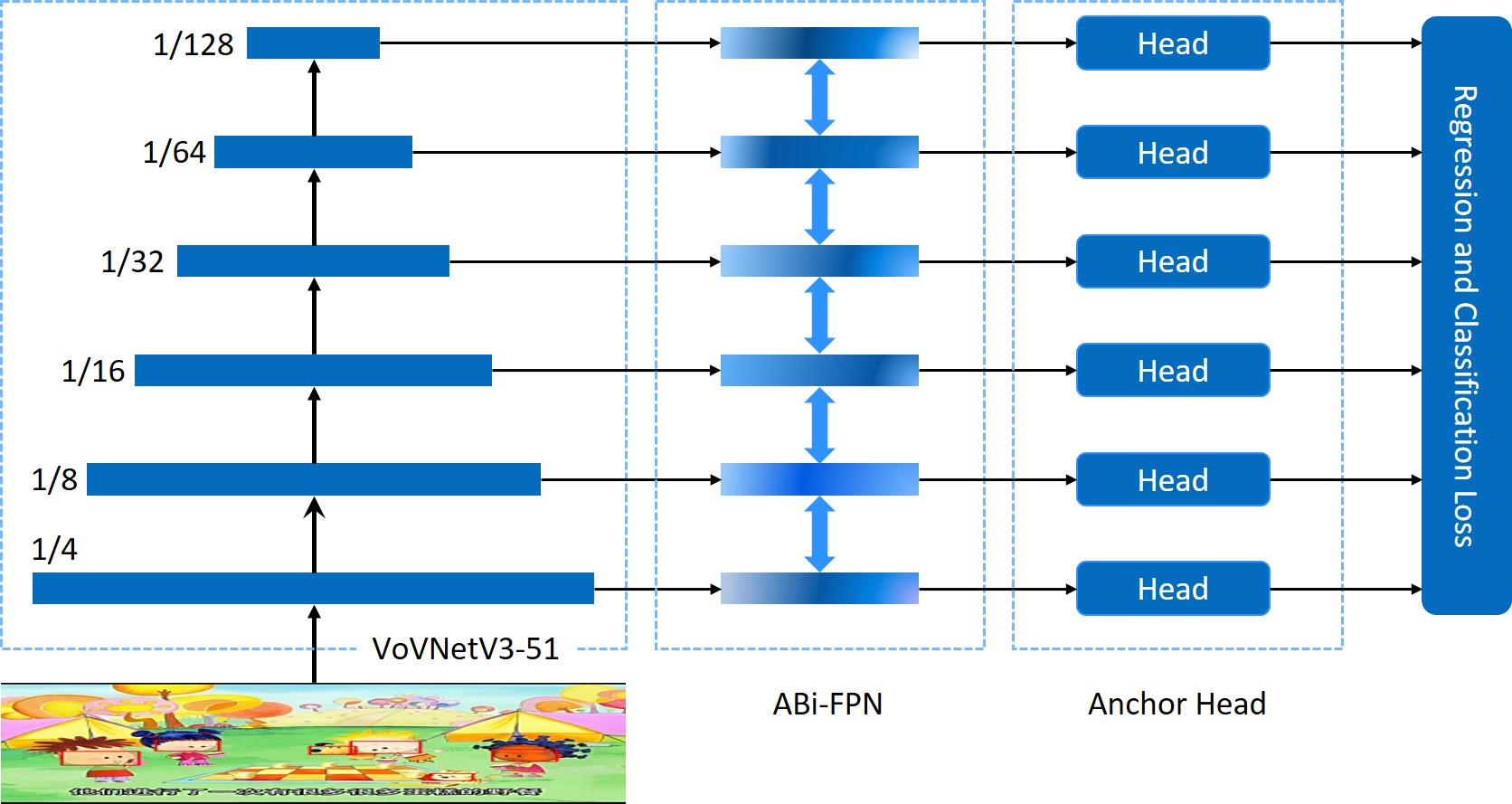}
    \vspace{-2mm}
    \caption{Pipeline of our ACFD, which is a anchor-based one-stage detector, where feature maps of backbone network with strides from $4$ to $128$ are feed into ABi-FPN and used for predicting.}
    \vspace{-3mm}
    \label{fig:pipeline}
\end{figure*}

To solve the aforementioned difficulties in cartoon faces, we propose a novel asymmetric cartoon face detector (ACFD) with four improvements to enhance the diversity of features, regression, and classification ability of networks according to the characteristics of cartoon faces. Specifically, to provide features with more diverse receptive fields, we propose a more effective backbone network VoVNetV3 based on VoVNet \cite{lee2019vovnet} and VoVNetV2 \cite{lee2020vovnetv2} with better performance than ResNet \cite{he2016resnet} and DenseNet \cite{huang2017densenet}, through which the multi-scale features can be extracted and then fused and enhanced by the following ABi-FPN. For enhancing the regression and classification capacity, we utilize DAM to compensate high-quality anchors with strong regression ability providing better initialization for the regressor, furthermore, MBC is used to better distinguish faces from the dense predictions. The main contributions of the paper can be summarized as follow.
\begin{itemize}
    \item Proposing a novel backbone network VoVNetV3 to obtain features with more diverse receptive fields.
    \item Designing an ABi-FPN to fuse multi-scale features and enhance the semantic information simultaneously.
    \item Presenting a DAM strategy to match high-quality anchors for each faces dynamically.
    \item Introducing a MBC module to distinguish faces within the dense predictions.
    \item Achieving the 1st place on the detection track of 2020 iCartoon Face Challenge.
\end{itemize}

\section{Related Work}
\paragraph{Face Detection.} Face detectors usually inherit the basic settings of generic object detection. Review the recent state-of-the-art face detectors \cite{zhang2017s3fd,tang2018pyramidbox,chi2019srn,li2019dsfd,zhang2020refineface}, all of them are anchor-based one-stage detectors and consist of backbone, neck and detection head. In detail, the backbone network designed for image classification \cite{he2016resnet,simonyan2014vgg} is used to extract multi-scale features, and the neck network is utilized to fuse and enhance the multi-scale features sequentially. We propose backbone network VoVNetV3 and ABi-FPN to extract multi-scale features with strong representation, aggregate, and enhance the context respectively.

\paragraph{Anchor Match.} Traditionally, an anchor would be assigned as positive if its Intersection over Union (IoU) with a ground-truth bounding box is larger than a predefined threshold \cite{ren2015fasterrcnn}. To suit face detection, \cite{zhang2017s3fd} proposes an improved strategy helping the outer faces to match enough anchors. \cite{chi2019srn} and \cite{zhang2020refineface} introduce selective two-step regression and classification to refine and filter the predefined anchors. \cite{kong2019consistent} and \cite{liu2019hambox} observe that taken regressed bounding boxes into match phase would bring considerable performance improvements. Inspired by these works, we propose a dynamic anchor match strategy to compensate enough high-quality anchors for the outer faces.

\paragraph{Loss Design.} Smooth-L1 and cross-entropy losses \cite{ren2015fasterrcnn} are widely used in object detection due to the simplicity and effectiveness. Nowadays, IoU-based losses \cite{rezatofighi2019giou,zheng2020diou} are proposed to strengthen the connections between regression and classification tasks, focal loss \cite{lin2017retinanet} is proposed to alleviate the extreme foreground-background class imbalance. Unlike these methods, we transfer the margin loss \cite{wang2018cosface,deng2019arcface} in face recognition to face detection for further improving the power of discrimination due to many cartoon faces are very similar to the background.

\section{Methodology}
In this section, we briefly introduce the pipeline of the proposed ACFD at first in Sec.~\ref{sec:pipeline}, then detailly describe the proposed VoVNetV3 in Sec.~\ref{sec:vovnetv3}, asymmetric bi-directional feature pyramid network in Sec.~\ref{sec:abi_fpn}, dynamic anchor match in Sec.~\ref{sec:dam}, and binary margin classification loss in Sec.~\ref{sec:mbc}, respectively.

\subsection{Pipeline of ACFD}\label{sec:pipeline}
The pipeline of the proposed ACFD is indicated in Fig.~\ref{fig:pipeline}. We adopt the proposed VoVNetV3-51 as the backbone network of ACFD, which consists of 6 stages to generate feature maps with strides from 4 to 128. Then, the multi-scale pyramid features extracted from the backbone network are feed into the proposed ABi-FPN for further aggregating and refining the context information. Finally, the dense predictions are obtained by the corresponding anchor head networks.

% \begin{figure*}[!t]
%     \centering
%     \includegraphics[width=0.95\linewidth]{./pic/VoVNetV3.jpg}
%     \caption{Architectures of (a) AOSA module, (b) and (c) ACB module for training and inference. $F_{1\times1}$ is convolution layer with kernel size $1$, $A_{3\times3}$ denotes ACB with kernel size 3, $F_{avg}$ is global average pooling, $W_C$ is fully-connected layer, $\oplus$ and $\otimes$ indicate element-wise addition and multiplication.}
%     \label{fig:vovnetv3}
% \end{figure*}
\begin{figure}[!htbp]
    \centering
    \includegraphics[width=0.95\linewidth]{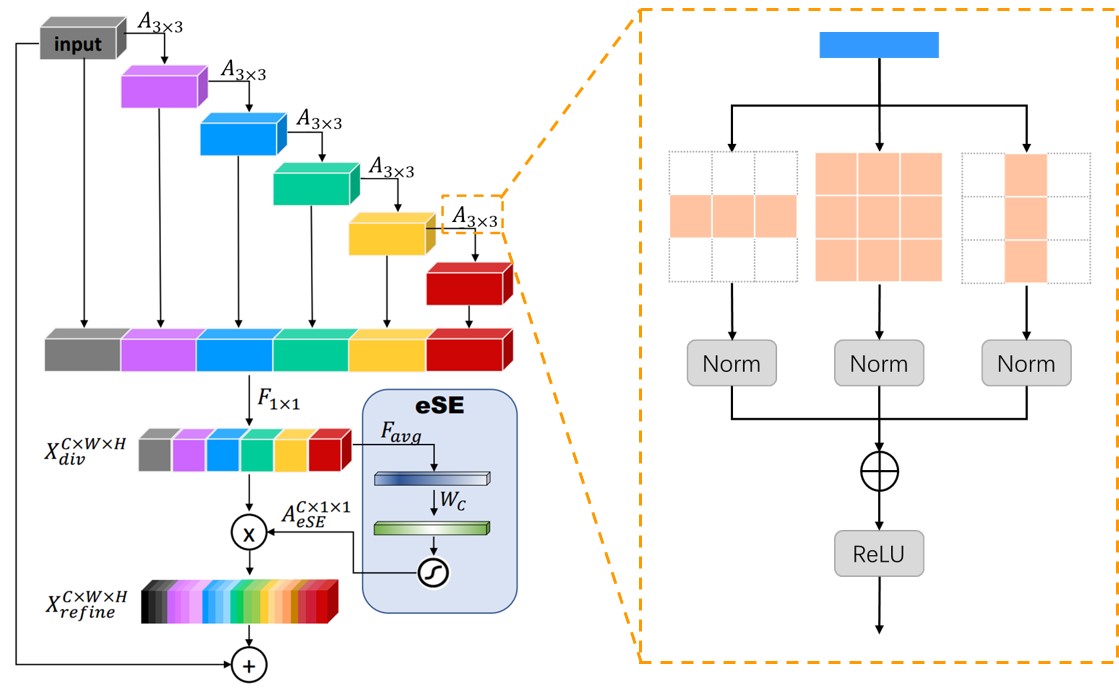}
    \vspace{-2mm}
    \caption{The architecture of AOSA module. $F_{1\times1}$ is convolution layer with kernel size $1$, $A_{3\times3}$ denotes ACB with kernel size 3, $F_{avg}$ is global average pooling, $W_C$ is fully-connected layer, $\oplus$ and $\otimes$ indicate element-wise addition and multiplication.}
    \label{fig:aosa}
    \vspace{-3mm}
\end{figure}
\subsection{VoVNetV3 Backbone Network}\label{sec:vovnetv3}
VoVNet \cite{lee2019vovnet} is a computation and energy efficient backbone network that can efficiently present diversified feature representations owing to the utilization of one-shot aggregation (OSA) modules. For further boosting the performance, VoVNetV2 \cite{lee2020vovnetv2} is proposed by adding the residual connection to address the limitation of VoVNet during optimization \cite{he2016resnet}, and employing an efficient squeeze-excitation (eSE) attention module for modeling the interdependency between the channel of feature maps to enhance its representation. In order to further enrich the diversity of feature maps, a more effective backbone network VoVNetV3 is proposed on the basis of VoVNet and VoVNetV2, which is comprised of several asymmetric one-shot aggregation (AOSA) modules, as presented in Fig.~\ref{fig:aosa}.

\paragraph{Asymmetric One-Shot Aggregation Module.} Unlike DenseNet \cite{huang2017densenet}, the OSA modules of VoVNet and VoVNetV2 generate feature maps in a relatively sparse-connected manner, where each feature map is connected to the subsequent convolution layer so as to produce the feature with a large receptive field, and concatenated with the final output feature map only once, receptively. As a result, OSA modules could generate features with rich receptive fields, however, both of them only process square receptive fields and would affect the detection of faces with different aspect ratios. Inspired by \cite{szegedy2017inception,ding2019acnet}, the proposed AOSA overcomes the above limitation by replacing all $3\times3$ convolutions in OSA with asymmetric convolution blocks (ACB) as shown in Fig.~\ref{fig:aosa}. In this way, two additional convolutions with kernel $1\times3$ and $3\times1$ are added in parallel on $3\times3$ convolution layer to extract features with rectangle receptive fields.

We utilize VoVNetV3-51 with 6 stages of outputs in this paper, its configuration is presented at Table. \ref{tab:config_of_vovnet51}.

\begin{table}[]
    \centering
    % \resizebox{\0.95\linewidth}{!}{
    \begin{tabular}{lcccc}
    \toprule
    Layer & Output & Stride & Repeat & Channel \\
    \toprule
    image & $640\!\times\!640$ & $-$ & $-$ & $-$ \\
    \midrule
    conv1 & $320\!\times\!320$ & $2$ & $1$ & $64$ \\
    conv2 & $320\!\times\!320$ & $1$ & $1$ & $64$ \\
    conv3 & $160\!\times\!160$ & $2$ & $1$ & $128$ \\
    \midrule
    stage1 & $160\!\times\!160$ & $1$ & $1$ & $128/256$ \\
    \midrule
    max-pool & $80\!\times\!80$ & $2$ & $1$ & $256$ \\
    stage2 & $80\!\times\!80$ & $1$ & $1$ & $160/512$ \\
    \midrule
    max-pool & $40\!\times\!40$ & $2$ & $1$ & $512$ \\
    stage3 & $40\!\times\!40$ & $1$ & $2$ & $192/768$ \\
    \midrule
    max-pool & $20\!\times\!20$ & $2$ & $1$ & $768$ \\
    stage4 & $20\!\times\!20$ & $1$ & $2$ & $224/1024$ \\
    \midrule
    max-pool & $10\!\times\!10$ & $2$ & $1$ & $1024$ \\
    stage5 & $10\!\times\!10$ & $1$ & $1$ & $128/128$ \\
    \midrule
    max-pool & $5\!\times\!5$ & $2$ & $1$ & $128$ \\
    stage6 & $5\!\times\!5$ & $1$ & $1$ & $128/128$ \\
    \bottomrule
    \end{tabular}
    \vspace{-2mm}
    \caption{The configuration of proposed VoVNetV3-51, in which the channel in format $-/-$ means the splited and concatenated channel in AOSA.}
    \vspace{-3mm}
    \label{tab:config_of_vovnet51}
\end{table}

\subsection{ABi-FPN}\label{sec:abi_fpn}
At present, most of the face detectors utilize ResNet \cite{he2016resnet} and VGG \cite{simonyan2014vgg} to extract the multi-scale features, while both of them can only possess square receptive fields and are harmful to the faces with extreme aspect ratios. It is especially important in cartoon face detection due to about $10\%$ of faces with ratios of larger than 2.0 or smaller than 0.5. As for solving the limitation of networks, recent state-of-the-arts face detectors \cite{tang2018pyramidbox,li2019dsfd,zhang2020refineface} add an additional module followed by feature fusion modules to refine the receptive fields, which is effective but inefficient. Instead of adding the extra modules for processing, we propose an efficient and effective module named ABi-FPN to fuse multi-scale features, enrich semantic information, and refine the receptive fields of features simultaneously. 
Specifically, ACB is also employed to replace the convolution layers of BiFPN \cite{tan2020efficientdet}, in which the receptive fields of features would be more diverse as the aggregation of multi-scale features.

\subsection{Dynamic Anchor Match (DAM)}\label{sec:dam}
\begin{figure}
    \centering
    \subfigure[]{
        \includegraphics[width=0.45\linewidth]{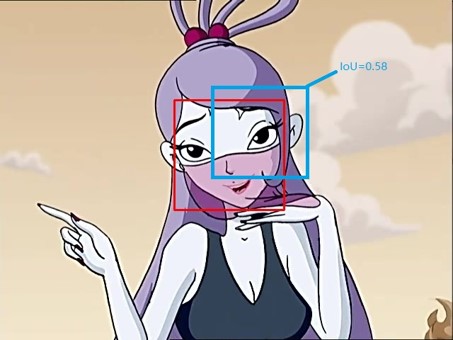}
    }
    \subfigure[]{
        \includegraphics[width=0.45\linewidth]{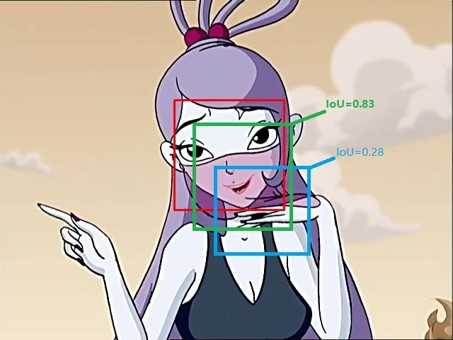}
    }
    \vspace{-5mm}
    \caption{Illustration of our DAM, where red boxes mean ground-truth faces, blue boxes are anchors, and green boxes are the corresponding regressed bounding boxes. (a) Assigning anchors with IoU greater than the predefined threshold as positive samples, (b) assigning unmatched anchors in the first step with strong regression ability as positive.}
    \vspace{-3mm}
    \label{fig:dynamic_anchor_match}
\end{figure}
Recently, some works \cite{kong2019consistent,liu2019hambox} observe an interesting phenomenon that some unmatched anchors have strong regression ability, as presented in Fig.~\ref{fig:dynamic_anchor_match} (b), the anchor with strong regression capacity could obtain a bounding box with the large IoU score, despite its own IoU is very small. Inspired by the observation, we propose a DAM strategy to make full use of these anchors with strong regression ability, as a result, it matches enough high-quality anchors for each ground-truth face. Firstly, the same as traditional match strategy \cite{liu2016ssd} shown as Fig.~\ref{fig:dynamic_anchor_match} (a), the anchors with IoU greater than a threshold are assigned as positive. Secondly, the anchors would be compensated as positive if IoUs of their related regressed bounding boxes are higher than another threshold. The details can be seen in Algorithm.~\ref{alg:dynamic_anchor_match}.

\vspace{-2mm}
\begin{algorithm}[!htbp]
\caption{Dynamic anchor match strategy.}\label{alg:dynamic_anchor_match}
\textbf{Input:} $A, B, T_1, T_2, G, R, L$\\
$A$ a set of predefined anchors, $B$ the related bounding boxes.\\
$T_1$, $T_2$ IoU thresholds for first and second match steps.\\
$G$ ground-truth boxes.\\
$R$, $L$ matched bounding boxes and labels of each anchor.\\
\textbf{Output:} $R, L$
\begin{algorithmic}[1]
    \FOR{$a_i$, $b_i$ in $A$, $B$}
    \STATE $\#$ first step
	\STATE $AnchorIoU_i\Leftarrow IoU(a_i, G)$
	\STATE $GtIdx, AnchorMax \Leftarrow argmax(AnchorIoU_i)$
	\IF{$AnchorMax \ge T_1$}
	\STATE $R[i] \Leftarrow G[GtIdx], L[i] \Leftarrow 1$
	\ELSE
	\STATE $\#$ second step
	\STATE $BboxIoU_i \Leftarrow IoU(b_i, G)$
	\STATE $GtIdx, BboxMax \Leftarrow argmax(BboxIoU_i)$
	\IF{$BboxMax \ge T_2$}
	\STATE $\#$ 2 means compesanted anchor
	\STATE $R[i] \Leftarrow G[GtIdx], L[i] \Leftarrow 2$
	\ELSE
	\STATE $L[i] \Leftarrow 0$
	\ENDIF
	\ENDIF
	\ENDFOR
\end{algorithmic}
\end{algorithm}
\vspace{-3mm}

\subsection{Margin Binary Classification Loss (MBC)}\label{sec:mbc}
As described in the above section, our DAM strategy could match enough high-quality anchors with strong regression ability for each face. However, these high-quality anchors are usually far from the ground-truth face and may dominate the loss during propagation. Therefore, we separately compute and weight the losses of matched anchor in the first step and compensated high-quality anchors in the second step, the regression and classification losses are reformatted as follow,
\begin{equation}
\vspace{-2mm}
    \ell_{reg} \!=\! \frac{1}{N_1}\!\sum_{i\in \psi_1}\!\mathcal{L}_{smoothL1}(x_i, x_i^{\ast}) \!+\! \frac{\lambda_{reg}}{N_2}\!\sum_{i\in \psi_2}\!\mathcal{L}_{smoothL1}(x_i, x_i^{\ast})
\end{equation}
\vspace{-6mm}
\begin{equation}
    \ell_{cls} \!=\! \frac{1}{N_1}\!\sum_{i\notin\psi_2}\!\mathcal{L}_{Focal}(p_i^m, p_i^{\ast}) \!+\! \frac{\lambda_{cls}}{N_2}\!\sum_{i\in \psi_2}\!\mathcal{L}_{Focal}(p_i^m, p_i^{\ast})
\end{equation}
where $N_1$ and $N_2$ are numbers of matched anchors $\psi_1$ and compensated anchors $\psi_2$, $\lambda_{reg}$ and $\lambda_{cls}$ are the corresponding weighted coefficients.

Furthermore, to improve the classification ability of our network so that it can distinguish faces who are similar to the background, we transfer the widely used margin-based loss function \cite{wang2018cosface,deng2019arcface} in face recognition to face detection. The margin-based losses share the same idea that maximizing the inter-class variance, minimizing the intra-class variance, and enhance the capacity of discrimination by adding an extra hard margin. In our margin binary classification application, suppose $p_i$ is the output of the network, then the margin-based prediction is formulated as follow,
\begin{equation}
    p_i^m = [p_i^o=1]\cdot(p_i\!-\!m) + [p_i^o=0]\cdot p_i
\end{equation}
in which, $p_i^m$ is the corresponding one-hot label, $m$ is a hard margin, and $p_i^m$ is used for the computation of classification loss.

\section{Experiments}
In this section, we firstly present the details of our implementation, then the effectiveness of our proposed ACFD is verified through a set of experiments, finally, we introduce the engineering tricks that are used for optimizing the time-consuming of our ACFD.
\subsection{Implementation Details}
% \vspace{-2mm}
\paragraph{Dataset.} During ablation studies, we split 50000 images of iCartoon Face \cite{li2019icartoonface} into 45000 images for training and 5000 for reporting the effectiveness. All images are used for training for the final submitted model.

% \vspace{-1mm}
\paragraph{Data Augmentation.} To prevent over-fitting and augment the difficult faces during training such as small and blur faces, the faces too large to location accurately, and etc. A series of data augmentations are employed, summarized as follow: (1) color distort for training images, (b) expand the images with a random range $[1, 4]$ by mean-padding to augment the small faces, (c) crop the images with a random size at a random position to augment the big faces, (d) random tile the faces to anchor scales, finally, resize the images to $640\!\times\!640$ for feeding into the network.

% \vspace{-1mm}
\paragraph{Anchor Design.} We associate only one anchor for each location of the detection layers with scale 4 and ratio $1\!:\!1$. So, there are total of 34125 anchors per image covered faces of size $16\!-\!512$ pixels within the training phase.

% \vspace{-1mm}
\paragraph{Optimization.} Without ImageNet-pretrained model, we use "kaiming" method \cite{he2015kaimingInit} to initialize all the parameters. The SGD algorithm is applied to train the model with momentum $0.9$, weight decay $5\!\times\!10^{-4}$, batch size $16\!\times\!4$ V100 GPUs, and warmup strategy to increase the learning rate from $1.0\!\times\!10^{-6}$ to $0.04$ at the first $1000$ iterations. Then, for the final submit, the learning rate is divided by $10$ at $200, 250, 280$ epochs, and ended at $300$ epoch; for the ablation studies, it is divided by 10 at $100, 150, 180$ epochs and ended at $200$ epoch. 

% \vspace{-1mm}
\paragraph{Other Hyper-parameters.} Empirically, we define two thresholds of dynamic anchor match as $T_1\!=\!0.35$ and $T_2\!=\!0.7$. Correspondingly, the weighted coefficients in regression and classification losses are $\lambda_{reg}\!=\!\lambda_{cls}\!=\!0.7$, and the margin used in classification loss is $0.2$.

% simulation: backbone
\begin{table*}[!htbp]
    \centering
    \begin{tabular}{l|ccccccc}
    \toprule
    Module & ResNet50 & SE-ResNet50 & Res2Net50 & ResNeSt50 & EfficientNet-B3 & VoVNetV2-39 & \textbf{VoVNetV3-51} \\
    \midrule
    mAP ($\%$) & $90.18$ & $90.23$ & $89.59$ & $89.97$ & $88.63$ & $90.37$ & $\mathbf{90.74}$ \\
    \bottomrule
    \end{tabular}
    \caption{mAP (\%) of ACFD with different backbone networks.}
    \label{tab:backbone}
    \vspace{-3mm}
\end{table*}

\vspace{-1mm}
\paragraph{Inference.} Multi-scale test is employed for the final submitting, we resize the images to three predefined scales that $480\!\times\!645$, $640\!\times\!860$ and $800\!\times\!1075$ for predicting. During inference, the network outputs the top-1000 predictions with confidence scores higher than 0.08 for each scale, finally, we apply the non-maximum suppression (NMS) algorithm with IoU threshold 0.55 to generate top-100 high confidence detections as the final results.

\subsection{Model Analysis}
In this subsection, extensive experiments are conducted to demonstrate the effectiveness of our proposed modules in ACFD. For the fair comparison, we use the same parameter settings apart from the specific changes to the components. In order to better understand our ACFD, the experiments are carried out on the different baselines, by ablating different modules to perform how it affects the final performance. 

\paragraph{VoVNetV3-51.} Firstly, we compare our VoVNetV3-51 with the state-of-the-art and widely used backbone networks, e.g., ResNet50 \cite{he2016resnet}, SE-ResNet50 \cite{hu2018senet}, Res2Net50 \cite{gao2019res2net}, ResNeSt50 \cite{zhang2020resnest}, EfficientNet-B3 \cite{tan2019efficientnet}, VoVNetV2-39 \cite{lee2020vovnetv2}. To better suit the framework of ACFD, two extra convolution layers with stride 2 followed by batch normalization and ReLU activation are added to these backbones for generating 6-level features with stride from 4 to 128. As presented in Table~\ref{tab:backbone}, ResNet50 and SE-ResNet50 perform the comparable performance on cartoon face detection with mAP $90.18$ and $90.23$, and far surpass the recent state-of-the-art Res2Net50, ResNeSt50, EfficientNet-B3 of $89.59$, $89.97$, $88.63$ respectively. Owing to the superiority of OSA module, VoVNetV2-39 achieves the better performance. By introducing asymmetric convolution blocks, the proposed VoVNetV3 achieves the highest mAP score $90.74$ since it can generate features with more diverse receptive fields, this is beneficial for the cartoon faces with large inter-class variances. 
%Although VoVNetV3 has more Parameters and FLOPs than others, they are introduced by the asymmetric convolution blocks and only involved in training phase, which can be merged during inference, see Sec.~\ref{sec:time_consuming} for details.

% \begin{table}[!htbp]
%     \centering
%     % \resizebox{0.98\linewidth}{!}{
%     \begin{tabular}{lccc}
%     \toprule
%     model & Parameters & FLOPs & mAP ($\%$) \\
%     \midrule
%     ResNet50 & 34M & & $90.18$ \\
%     SE-ResNet50 & & & $90.23$ \\
%     Res2Net50 & & & $89.59$ \\
%     ResNeSt50 & & & $89.97$ \\
%     EfficientNet-B3 & & & $88.63$ \\
%     VoVNetV2-39 & & & $90.37$ \\
%     \textbf{VoVNetV3-51} & & & $\mathbf{90.74}$ \\
%     \bottomrule
%     \end{tabular}
%     \caption{mAP (\%) of different backbone networks, in which Parameters and FLOPs are measured with a $640\!\times640$ image.}
%     \label{tab:backbone}
%     \vspace{-3mm}
% \end{table}

\paragraph{A-BiFPN.} Next, take ACFD with ResNet50 backbone as the baseline, a series of simulations are carried to verify the effectiveness of proposed ABi-FPN by comparing with the plain FPN \cite{lin2017fpn}, BiFPN \cite{tan2020efficientdet} and SEPC \cite{wang2020SEPC}. As presented in Table~\ref{tab:fpn}, benefit from the ability of BiFPN to aggregate multi-scale feature maps by top-down and bottom-up paths, it outperforms the plain FPN and SEPC by $1.8\%$ and $1.3\%$ points. The proposed ABi-FPN further enhances this ability by employing asymmetric convolution blocks to capture the features with more diverse receptive fields, and achieves the better mAP score of $90.36\%$.

\begin{table}[!htbp]
    \centering
    % \resizebox{0.9\linewidth}{!}{
    \begin{tabular}{l|cccc}
    \toprule
    Module & FPN & BiFPN & SEPC & \textbf{ABi-FPN} \\
    \midrule
    mAP ($\%$) & $88.30$ & $90.18$ & $88.80$ & $\mathbf{90.36}$ \\
    \bottomrule
    \end{tabular}
    \caption{mAP ($\%$) of different feature pyramid networks on ACFD.}
    \label{tab:fpn}
    \vspace{-3mm}
\end{table}

\paragraph{Dynamic Anchor Match.}
Then, several experiments are conducted to evaluate the superiority of the proposed DAM, as shown in Table~\ref{tab:dam}. We select the model with ResNet50 backbone, the plain FPN and classic match strategy as the baseline, it gets mAP of $87.65\%$. In particular, the worse performance is obtained while compensating anchors by using regressed bounding boxes with a lower IoU threshold $T_2\!=\!0.35$, this is due to many low-quality anchors are matched for regression causing too many false-positive results. On the other hand, when compensating anchors with a higher IoU threshold but summing the losses by weights $\lambda_{reg}\!=\!\lambda_{cls}\!=\!1.0$, the performance is not the best due to the loss of compensated anchors may dominate. By choosing $T_1\!=\!0.35, T_2\!=\!0.7, \lambda_{reg}\!=\!\lambda_{cls}\!=\!0.7$, the model equipped with DAM achieves mAP score of $88.90\%$, it is about $1.3\%$ points higher than the baseline.

\begin{table}[!htbp]
    \centering
    \begin{tabular}{cccc|c}
    \toprule
    $T_1$ & $T_2$ & $\lambda_{res}$ & $\lambda_{cls}$ & mAP ($\%$) \\
    \midrule
    - & - & - & - & $87.65$\\
    $0.35$ & $0.35$ & $1.0$ & $1.0$ & $87.56$\\
    $0.35$ & $0.7$ & $1.0$ & $1.0$ & $88.60$\\
    $0.35$ & $0.35$ & $0.5$ & $0.5$ & $87.54$\\
    $0.35$ & $0.7$ & $0.5$ & $0.5$ & $88.75$\\
    $\mathbf{0.35}$ & $\mathbf{0.7}$ & $\mathbf{0.7}$ & $\mathbf{0.7}$ & $\mathbf{88.90}$\\
    \bottomrule
    \end{tabular}
    \caption{mAP ($\%$) of ACFD matching anchors by proposed DAM with different parameters.}
    \label{tab:dam}
    \vspace{-3mm}
\end{table}

% Visualization
\begin{figure*}[!htbp]
    \vspace{-2mm}
    \centering
    \includegraphics[width=0.95\linewidth]{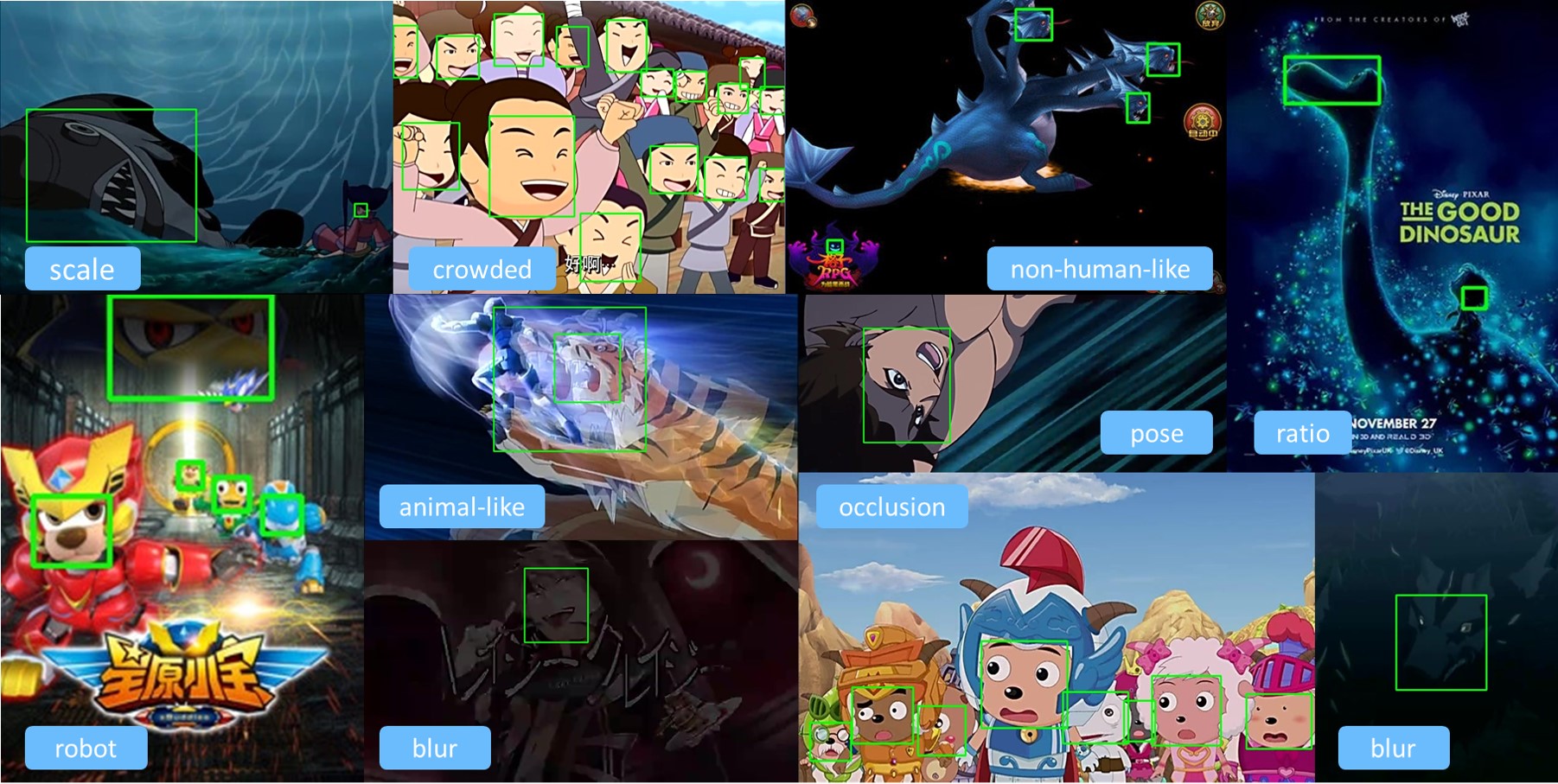}
    \caption{Illustration of our ACFD to various large variations on iCartoonFace dataset. Green buonding boxes indicate the detector confidence is above 0.7.}
    \vspace{-4mm}
    \label{fig:visual}
\end{figure*}

\paragraph{Margin Binary Classification.} At the next stage, the effectiveness of different margins in MBC is verified by simulations shown in Table~\ref{tab:mbc}, in which $m\!=\!0$ means margin loss is disabled. Too small margin value makes the model not work and too large value makes the models hard to optimized, $m\!=\!0.2$ is adopted in our ACFD.

\begin{table}[!htbp]
    \centering
    \begin{tabular}{l|ccccc}
    \toprule
    margin ($m$) & $0$ & $0.1$ & $\mathbf{0.2}$ & $0.3$ & $0.5$ \\
    \midrule
    mAP ($\%$) & $90.48$ & $90.51$ & $\mathbf{90.73}$ & $90.67$ & $90.41$ \\
    \bottomrule
    \end{tabular}
    \caption{mAP ($\%$) of ACFD with different margin parameters.}
    \label{tab:mbc}
    \vspace{-3mm}
\end{table}

\paragraph{Ablation Study.} Finally, we take model with ResNet50 backbone and the plain FPN as the baseline and demonstrate the effectiveness of proposed modules by adding each of them to the baseline. As listed in the first and second rows of Table~\ref{tab:ablation_study}, the baseline achieves mAP of $87.85\%$ and $87.13\%$ with and without the multi-scale test. By adding all modules to the baseline, mAP scores $90.94\%$ and $90.33\%$ are obtained with and without the multi-scale test, which surpasses the baseline more than $3.0$ points mAP. This final model achieves $92.91\%$ on the leaderboard of competition and wins the first place with a large margin. Fig.~\ref{fig:visual} shows more examples to demonstrate the effectiveness of ACFD.

\begin{table}[!htbp]
    \centering
    \resizebox{0.98\linewidth}{!}{
    \begin{tabular}{ccccc|c}
    \toprule
    VoVNetV3 & ABi-FPN & DAM & MBC & ms-test & mAP ($\%$) \\
    \midrule
    & & & & & $87.13$ \\
    & & & & $\checkmark$ & $87.85$ \\
    & & $\checkmark$ & & & $88.47$ \\
    $\checkmark$ & & $\checkmark$ & & & $89.46$ \\ 
    & $\checkmark$ & $\checkmark$ & & & $89.39$ \\
    $\checkmark$ & $\checkmark$ & $\checkmark$ & & & $90.17$ \\
    $\checkmark$ & $\checkmark$ & $\checkmark$ & $\checkmark$ & & $90.33$ \\
    $\checkmark$ & $\checkmark$ & $\checkmark$ & $\checkmark$ & $\checkmark$ & $\mathbf{90.94}$ \\
    \bottomrule
    \end{tabular}}
    \vspace{-2mm}
    \caption{Effectiveness of our proposed modules. Taken a model with ResNet50 backbone as baseline, all models are trained on 45000 training images and evaluated with mAP ($\%$) on 5000 validating images.}
    \label{tab:ablation_study}
    \vspace{-5mm}
\end{table}

\subsection{Time-Consuming Optimization}\label{sec:time_consuming}
\begin{figure}
    \centering
    \includegraphics[width=0.95\linewidth]{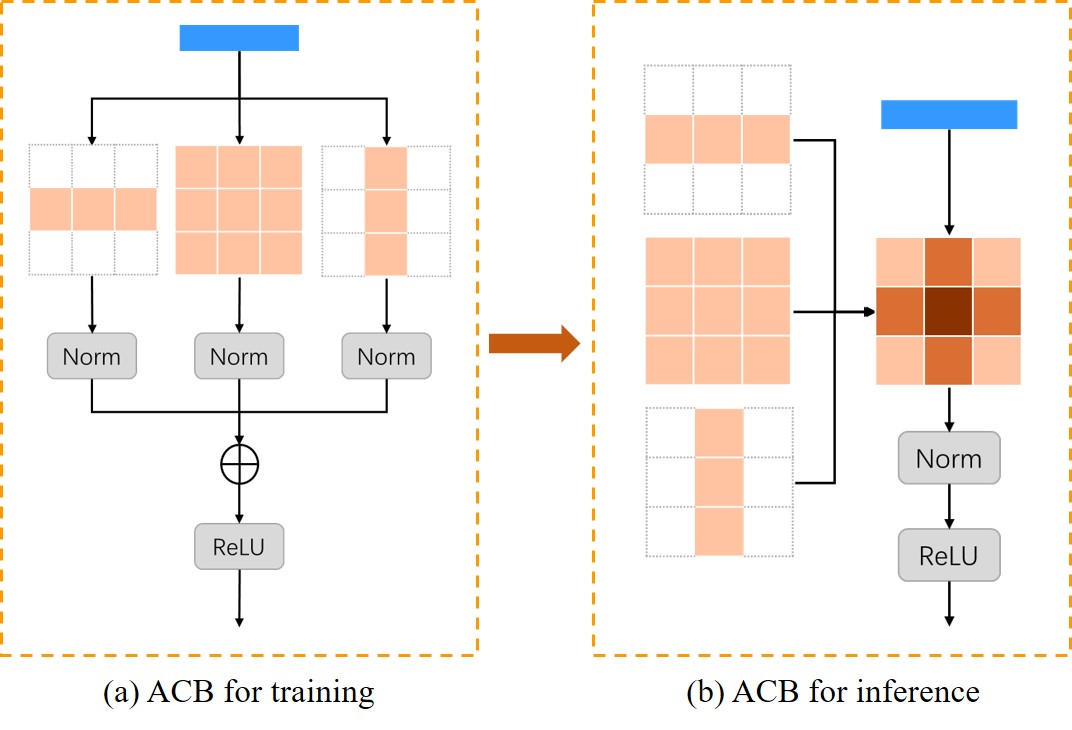}
    \vspace{-4mm}
    \caption{Illustration of asymmetric convolution blocks (a) original structure for training and (b) merged three convolutions into one for inference.}
    \vspace{-5mm}
    \label{fig:acb}
\end{figure}
In order to meet the competition requirements that the inference time for a single image should not exceed 50 ms, we utilize some engineering tricks to optimize our ACFD. Firstly, due to the convolution and normalization operations in asymmetric convolution blocks are linear, so the horizontal and vertical branches can be merge to become a classic convolution block, as shown in Fig.~\ref{fig:acb}. Furthermore, the convolution and normalization operations can be merged into a single convolution operation as follow,
\begin{align}
	a=\frac{\gamma}{\sqrt{\delta^2+\epsilon}},W'=W\ast a,B'=(B-\mu)\ast a + \beta,
\end{align}
where $\mu$, $\delta$, $\gamma$, and $\beta$ are mean, variance and affine parameters of origin normalization operation, $W$, $B$ and $W'$, $B'$ are parameters of convolution layer before and after merging. Finally, we convert the PyTorch model to the TensorRT version by torch2trt\footnote{https://github.com/z-bingo/torch2trt} tool and evaluate the model with a large batch size to further accelerate the inference speed.

\vspace{-2mm}
\section{Conclusion}
Aiming at the difficulties of cartoon face detection, a novel asymmetric cartoon face detector (ACFD) is proposed in this paper. In particular, a strong backbone network VoVNetV3 is introduced to extract feature maps with more diverse receptive fields benefit from the asymmetric convolution blocks; then, the multi-scale features are better aggregated and enhanced by the proposed ABi-FPN. In addition, we propose a dynamic anchor match strategy to match high-quality anchors for each face by making full use of the regressed bounding boxes; furthermore, margin binary classification loss is used to enhance the discrimination of network to better distinguish faces from the dense predictions.

\clearpage
%% The file named.bst is a bibliography style file for BibTeX 0.99c
\bibliographystyle{named}
\bibliography{ACFD}

\end{document}